\documentclass[sigconf,nonacm]{acmart}
\AtBeginDocument{%
  \providecommand\BibTeX{{%
    \normalfont B\kern-0.5em{\scshape i\kern-0.25em b}\kern-0.8em\TeX}}}

\setcopyright{acmcopyright}
\copyrightyear{2023}
\acmYear{2023}
\acmDOI{XXXXXXX.XXXXXXX}

\acmConference[CIKM'23]{32nd ACM International Conference on Information and Knowledge Management}{21-25 October, 2023}{Birmingham, UK}
\acmBooktitle{CIKM'23: 32nd ACM International Conference on Information and Knowledge Management, 21th - 25th October, 2023} 
\acmPrice{15.00}
\acmISBN{978-1-4503-XXXX-X/18/06}

\usepackage[british]{babel}

\begin{document}

\title{Product Information Extraction using ChatGPT}

\author{Alexander Brinkmann}
\email{alexander.brinkmann@uni-mannheim.de}
\affiliation{%
  \institution{University of Mannheim}
  \city{Mannheim}
  \country{Germany}
}

\author{Roee Shraga}
\email{r.shraga@northeastern.edu}
% \authornotemark[2]
\affiliation{%
  \institution{Northeastern University}
  \city{Boston}
  \state{MA}
  \country{USA}
}

\author{Reng Chiz Der}
\email{rder@mail.uni-mannheim.de}
\affiliation{%
  \institution{University of Mannheim}
  \city{Mannheim}
  \country{Germany}
}

\author{Christian Bizer}
\email{christian.bizer@uni-mannheim.de}
\affiliation{%
  \institution{University of Mannheim}
  \city{Mannheim}
  \country{Germany}
}

\renewcommand{\shortauthors}{Brinkmann et al.}

\begin{abstract}
Structured product data in the form of attribute/value pairs is the foundation of many e-commerce applications such as faceted product search, product comparison, and product recommendation. Product offers often only contain textual descriptions of the product attributes in the form of titles or free text. Hence, extracting attribute/value pairs from textual product descriptions is an essential enabler for e-commerce applications.
In order to excel, state-of-the-art product information extraction methods require large quantities of task-specific training data. The methods also struggle with generalizing to out-of-distribution attributes and attribute values that were not a part of the training data.
 Due to being pre-trained on huge amounts of text as well as due to emergent effects resulting from the model size, Large Language Models like ChatGPT have the potential to address both of these shortcomings. 
This paper explores the potential of ChatGPT for extracting attribute/value pairs from product descriptions. We experiment with different zero-shot and few-shot prompt designs.
Our results show that ChatGPT achieves a performance similar to a pre-trained language model but requires much smaller amounts of training data and computation for fine-tuning.
%Our results show that adding three related demonstrations improves the results by 7\% in F1 score.
%\cb{Todo: Change last sentense to state that LLMs achieve a similar performance requireing a much smaller amount of training data compared to PLMs.}
\end{abstract}




\maketitle

\section{Introduction}
\label{sec:introduction}

Product attribute/value pairs are crucial for e-commerce platforms since they enable online shoppers to use faceted product search~\cite{wei_survey_2013} and to compare products along explicit criteria. 
On e-commerce marketplaces and open product catalogues, product attribute/value pairs are often missing because merchants only provide unstructured product information, such as titles and descriptions~\cite{yang_mave_2022,zheng_opentag_2018}.
%Todo Alex (Done): Explain in which context they are noisy and have missing values? In product catalogs? Which and why? Maybe it is better to argue via large open marketplaces or plattforms on which such attributes are not avalable as merchants just submit textial product descriptions.
Product information extraction extracts attribute/value pairs from these product titles and product descriptions.
%Todo Alex (Done):  Same here: Do not use the terms "missing" and "recovering" but a proper definition of the term.
Existing works can be distinguished into closed extraction and open extraction approaches. Closed extraction assumes that the relevant attributes are known prior to the extraction ~\cite{yang_mave_2022, zheng_opentag_2018, zhu_multimodal_2020, xu_scaling_2019, wang_learning_2020}. Open extraction assumes that neither a complete attribute set nor all attribute values are known prior to the extraction~\cite{zhang_oa-mine_2022}. 
% Todo Alex (Done): Please use a definition that covers open and closed extraction as well, not only closed.
For example, consider the following title of a product offer \emph{``Canon EOS 1000D SLR 10 MP EF-S 18-55''}. A closed extraction approach extracts from this title an explicit attribute like \emph{``Resolution``} and returns the attribute/value pair \emph{``Resolution: 10 MP``}. An open extraction extracts all attribute/value pairs and returns a list of attribute/value pairs like \emph{``Brand: Canon``, ``Megapixels: 10``}.
%Can be removed as you also say this later. In this paper, we systematically evaluate how ChatGPT extracts attribute/value pairs from product titles. For this purpose, we experiment with closed and open prompts for product information extraction. 
%, because product titles are carefully designed by retailers~\cite{xu_scaling_2019} and consequently form a valuable source for missing attribute/value pairs.
%  Todo Alex (Done): "A user willing to extract" is not a realistic use case, refine.
%a user willing to extract a \emph{``Resolution``} attribute for the product. Closed product information extract aims at extracting the attribute/value pair \emph{``Resolution: 10 MP``} from the title.

% Todo Alex (Done): Do not use "we focus" as this is vague. Be concrete in what the paper does.

% Todo Alex (Done): Say "state of the art" models often rely on PMLs or provide a bit deeper overivew of history of product data extraction methods.
\begin{sloppypar}
Many state-of-the-art product information extraction methods~\cite{zhu_multimodal_2020, xu_scaling_2019, yang_mave_2022, zheng_opentag_2018}
%Todo Alex: Please cite some of these methods here!
rely on pre-trained language models (PLM) such as BERT.
The different approaches encode a product title using the PLM and add additional layers on top, which tag the sequence of an attribute value in the title.
% Todo Alex (Done): Also cover the open use case here. So no "requested"
OA-Mine is an example of an open extraction approach that relies on a PLM to generate attribute value candidates from product titles and to discover new attributes from the generated attribute value candidates~\cite{zhang_oa-mine_2022}.
The approaches have two main drawbacks: (1) they require training data to fine-tune the models, and (2) the fine-tuned models have problems generalizing to unseen attributes and attribute values.
% Todo Alex (Done): Also say that they have problems generalizing to unseen (or use term out-of-distribution) attributes (not only attribute values).
\end{sloppypar}

%Recent work has shown that scaling up the number of parameters and the training compute of language models can overcome drawbacks of models like BERT. 
Large autoregressive language models (LLMs) such as ChatGPT~\cite{ouyang_training_2022}, BLOOM~\cite{yong_bloom1_2023} or PaLM~\cite{chowdhery_palm_2022} have recently shown their potential to overcome these shortcomings. Due to being pre-trained on huge amounts of text as well as due to emergent effects resulting from the model size~\cite{wei_emergent_2022}, LLMs often demonstrate a better zero-shot performance compared to PLMs like BERT and are more robust to unseen examples~\cite{brown_language_2020}.

In this paper, we evaluate how ChatGPT extracts attribute/value pairs from product titles using prompt engineering and in-context learning
%Todo Alex (Done): Emergent abilities only refers to ability to do stuff that is not mentioned in training data. Here you need to also mention large training corpora.
and make the following contributions:
\begin{itemize}
    \item We evaluate the performance of ChatGPT (gpt3.5-turbo-0301) for closed and open product information extraction.
    \item We systematically explore different prompt designs and show the performance of the best zero-shot prompt design is similar to a fine-tuned pre-trained language model.
    \item We experiment with in-context learning for product information extraction and show that providing only three demonstrations is sufficient to reach an F1 score of 94.51\%.
    %\item We analyze the extracted attribute/value pairs and show how ChatGPT normalizes the attribute/value pairs.
    %Todo Alex: Remove the last contribution if you do not have space in the paper to propoerly present this.
\end{itemize}

This paper is structured as follows. Section \ref{sec:experimental_setup} introduces the dataset and the experimental setup. Section \ref{sec:prompt_design} and Section \ref{sec:in_context_learning} explain different prompt designs, in-context learning and present results. The code and the dataset to reproduce our results are online available\footnote{https://github.com/wbsg-uni-mannheim/pie\_chatgpt}.

\section{Experimental Setup}
\label{sec:experimental_setup}

In this section, we introduce our experimental setup. For this setup, we discuss the MAVE dataset and how we prepared it for our experiments, the API calls against the OpenAI API, the evaluation and the baselines.

% Todo Alex (Done): Please improve description of MAVE dataset. The current version is not very clear and mixes review and product description related terminology. Please also mention size (attributes and examples) of the dataset to lay the foundation for next section.
\vspace{.2cm}\noindent\textbf{Mave Dataset.}
We experimented with a subset of product titles from the MAVE dataset~\cite{yang_mave_2022}. MAVE is a product dataset for multi-source attribute value extraction. MAVE is derived from a public collection of product offers from the e-commerce platform Amazon~\cite{ni_justifying_2019}. The authors of MAVE add annotations of attribute/value pairs for attribute values that can be found in the product titles and descriptions. To obtain these annotations, human experts define a set of categories and a list of attributes that are relevant for each category. Afterwards, each product offer is assigned to a product category and category-specific attribute value spans are annotated in the titles and descriptions of the product offers. An ensemble of five differently fine-tuned versions of the AVEQA~\cite{wang_learning_2020} model is used for this annotation. Each final record contains a single product offer with a product title, a product description, an attribute and annotation information about attribute values in the title or the description. Each record is attribute-dependent because a product offer can appear multiple times with different product attributes. The MAVE dataset contains in total 3.5 million unique products, 4.8 million unique attribute/value pairs, 2.4 thousand unique categories and 1.4 thousand unique attributes.

\vspace{.2cm}\noindent\textbf{Preparing the Dataset for ChatGPT.}
In order to keep the costs of experimenting with the OpenAI API manageable, we restrict our experiments to a subset of product offers.
The product offers are drawn from three selected product categories and have at least one annotation for one of 15 selected attributes. The selected categories and attributes are listed in Table \ref{tab:categories_attributes}.%.
% Table \ref{tab:categories_attributes} lists the selected product categories and product attributes.

% \usepackage{booktabs}
\begin{table}[h]
\caption{Categories and attributes per category from the MAVE dataset selected for our experiments.}
\label{tab:categories_attributes}
\begin{tabular}{@{}ll@{}}
\toprule
\textbf{Category}           & \textbf{Attribute}     \\ \midrule
Digital Camera     & Camera Weight, Optical Zoom,        \\ 
                   &  Resolution, Sensor Size, Sensor Type   \\ \midrule
Memory Cards &  Capacity , SD Format                          \\ \midrule
Laptops            & Battery Life, No. Cores, Processor Brand,      \\
                   &  Processor Speed, Refresh Rate, Resolution,  \\
                   & Screen Size, Weight \\ \bottomrule
\end{tabular}
\end{table}

%Todo Alex (Done):  Add a complete list of the attributes to the paper, so that the reader better understands the task and can judge the difficulty. Similar to Table 2 in Ket's paper: https://arxiv.org/pdf/2306.00745.pdf

We split the dataset into train:eval:test=8:1:1 using MAVE's shared code. To prevent data leakage we made sure that each product is contained either in train, validation and test. After splitting the training set contains 71,266 records and the validation set contains 8,789 records.
For the final test set, we selected up to 40 product titles for each category and attribute combinations  
%Todo (DOne) Alex: Remove term "requested value" from the paper. You do this for all attribtues in your set of 15 attributes, right?
for which the attribute value is contained in the product title and up to 10 product titles for which the attribute value is not contained in the product title. In total, the test set contains 562 test records.

During our experiments, we realized that ChatGPT normalizes attribute values. Table \ref{tab:value_normalization} shows different values annotated in the MAVE dataset and how ChatGPT normalizes these values.
%For example the attribute value \emph{``8 hour``} of the attribute \emph{``Battery Life``} is normalized to \emph{``8 hours``} and the attribute value \emph{``13.3-inch``} of the attribute \emph{``Screen Size``} is normalized to \emph{``13.3 inches``}. 
The normalizations improve the quality of the attribute/value pairs. Therefore, we manually verified the existing attribute value annotations and added normalized attribute values to the ground truth if applicable. We added normalized annotations to 88.65\% of the MAVE annotations in the test set. We share the test set in our GitHub repository.
%During our experiments, we realized that the target attribute values provided by MAVE are not normalized.The reason is that MAVE annotates the longest detected sequence of the initially used ensemble model. Therefore, all target attribute values in the test set are manually verified and updated by a human annotator. For example, for the original target attribute values "Intel Core i3" and "Intel Core" a unified value "Intel" is added if the goal is to extract the processor brand of a laptop. 

\begin{table}[h]
\caption{Annotated values in MAVE and how ChatGPT normalizes these values for different categories and attributes.}
\label{tab:value_normalization}
\begin{tabular}{@{}llll@{}}
\toprule
Category & Attribute         & Annotated & Normalized  \\
 &          & MAVE Value  & ChatGPT Value \\ \midrule
Digital Camera & Resolution   & 2 Megapixel              & 2MP                         \\
Laptops & No. Cores     & Dual-Core                & 2                           \\
Laptops & Battery Life        & 8 Hour                   & 8 Hours                     \\
Laptops & Screen Size         & 13.3-Inch                & 13.3 inches                 \\
Memory Cards & Capacity & 8 Gigabytes              & 8GB                         \\ \bottomrule
\end{tabular}
\end{table}

%Todo: You need to be more precise here. What do you do? What do you mean with extension?

\vspace{.1cm}\noindent\textbf{API Calls.}
The down-sampled version of the MAVE dataset contains 562 test records, which corresponds to the number of API calls per experiment. We use the ChatGPT version \emph{(gpt3.5-turbo-0301)} for all experiments and set the temperature parameter to 0 to make experiments reproducible. We use the langchain\footnote{https://python.langchain.com/en/latest/index.html} python package to call the API and track the number of tokens used by the model because the number of used tokens dictates the cost of the model usage\footnote{\$0.002 per 1,000 tokens. See https://openai.com/pricing}.

\vspace{.2cm}\noindent\textbf{Evaluation.}
The model responds with natural language texts. To decide if the target attribute value corresponds to the model's answer, we strip off whitespaces before and after the answer and check if the processed text exactly matches the target attribute value of the ground truth.
Following previous works~\cite{yang_mave_2022, yan_adatag_2021, xu_scaling_2019, zhang_oa-mine_2022}, we use Precision, Recall and F1 score as evaluation measures. Additionally, we track the average cost per request.

\vspace{.2cm}\noindent\textbf{Baselines.}
%Todo Alex: Be more precise here: What does "inspired" mean? Clearly state how the approach that you use related to OpenTag and ADOpenTag.
We train two closed extraction models as baselines for our experiments. Both baseline models rely on a PLM.
The first baseline is a fully-trained AVEQA~\cite{wang_learning_2020} model, using 71,266 training records. AVEQA formulates the attribute/value pair extraction as a question-answering task. The model encodes both attribute and product title with a BERT encoder and predicts the attribute value. The model is fine-tuned for 20 epochs on the reduced MAVE training set with a learning rate of 3e-5 and a batch size of 32.
The second baseline is a named entity recognition model (NER) for product attribute extraction~\cite{putthividhya_bootstrapped_2011}. For this NER model, we group the training set by product offer resulting in 18,083 records for training. DeBERTa is used to tokenize and encode the product titles~\cite {he_deberta_2021}. The tokens are annotated such that the model learns to classify tokens that describe an attribute value as mentioned in the training set. The model is fine-tuned for 10 epochs on the reduced MAVE training set with a learning rate of 2e-5 and a batch size of 8. Note that both baselines are fully trained over thousands of samples and their training took hours ($\sim$10 hrs. for AVEQA and $\sim$1.5 hrs. for NER).

%and annotate all tokens in the product title that mention an attribute value. The source for the annotations is the evidence list provided by MAVE. The model learns to classify the tokens according to the annotation.

%Our baseline is inspired by OpenTag~\cite{zheng_opentag_2018} and ADOpenTag~\cite{yang_mave_2022}. OpenTag uses a BiLSTM-Attention-CRF architecture for named entity recognition. We replace the BiLSTM-Attention-CRF architecture with a state-of-the-art DeBERTa model because this model has been successful on the . OpenTag learns one model per attribute. 
%Todo Alex: Unclear what you do here and why and why the result still related to OpenTag. You basicially use a different approach, right? Then you need to present this differently. e.g. say that you use a simple PML-based method as baseline and do not relate your baseline to OpenTag which does something different.
%In order to learn one model for multiple attributes, we concatenate product title, category and attribute name to make the input to the model attribute dependent. We refer to this baseline as ADOpenTag-DeBERTa. ADOpenTag is finetuned for 5 epochs with a learning rate of 2e-5 on the training split of our downsampled dataset.

%We trained two baselines on our reduced training set of the MAVE dataset.

\begin{figure*}[ht]
\includegraphics[width=\textwidth]{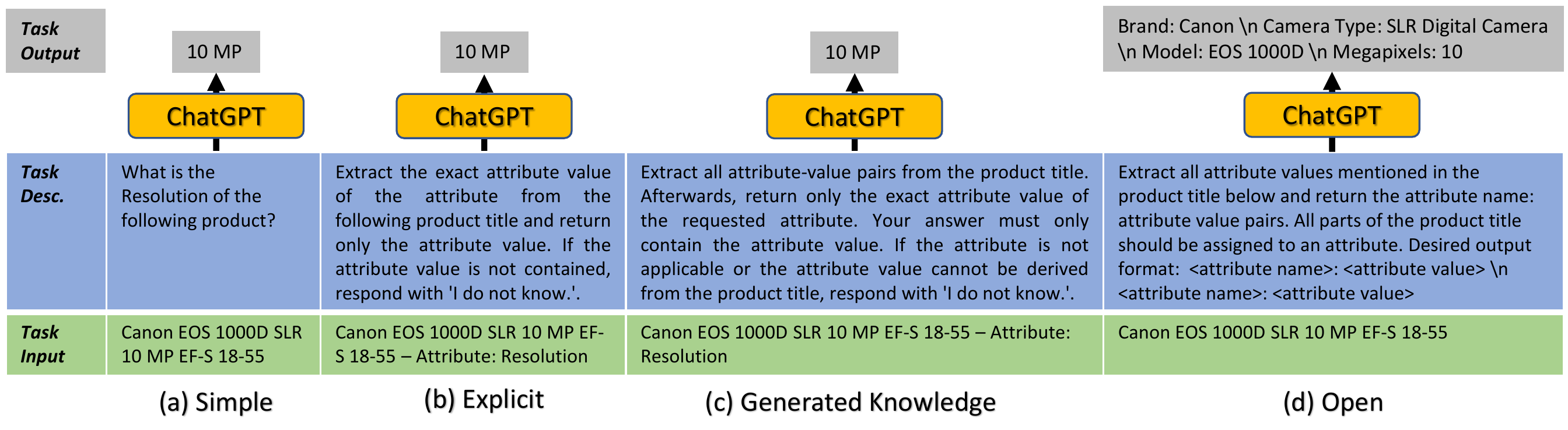}
\caption{Zero-shot prompt designs: Prompts (a), (b) and (c) are examples of closed extraction corresponding to Simple, Explicit and Generated Knowledge prompt designs. Prompt (d) is an example of open information extraction.}
\label{fig:extraction_prompt_new}
\end{figure*}
\section{Zero-shot Prompt Design}
\label{sec:prompt_design}

LLMs like ChatGPT expect text as input and return text as output.
The input text is also known as a prompt. In our case, the prompt contains a task description and a task input.
Recent work has shown that careful prompt design is crucial to achieving good performance~\cite{narayan_can_2022, zhao_calibrate_2021}.
In this section, we experiment with different prompt designs to explore the zero-shot performance of ChatGPT. The prompt designs for product information extraction can be categorized into closed and open.
We first introduce the closed and open prompt designs and afterwards present results.

%\subsection{Closed Information Extraction}
%\label{sub-sec:closed_information_extraction}
\vspace{.2cm}\noindent\textbf{Closed Product Information Extraction.}
The goal of our closed prompts is to extract the value of a specific attribute from a product title. % Changed from product title to desctiption as we are talking about the general case here.
The task descriptions of closed product information prompts are categorized as follows:

\begin{itemize}
    \item \textbf{Simple}: Task description asks for an attribute using simple language. 
    \item \textbf{Request}: Task description requests a specific attribute. %\rs{Explicit or Request?} \rs{Explicit is a comb.}
    \item \textbf{Short answer (sh.)}: Task description asks the model to respond only with the attribute value.
    \item \textbf{Unknown (unk.)}: Task description tells the model how to answer if the attribute value cannot be derived.
    \item \textbf{Generated Knowledge}: Task description asks the model to first implicitly extract all attribute-value pairs before extracting an explicit attribute~\cite{liu_generated_2022}. 
\end{itemize}

We refer to a prompt that combines request, short and unknown as \textbf{Explicit}. The prompts (a), (b) and (c) in Figure \ref{fig:extraction_prompt_new} are example prompts for closed extraction: (a) is a simple prompt, (b) is an explicit prompt, %that combines the ideas of request, simple and unknown in its task description, 
and (c) is a prompt that uses generated knowledge.

%Todo Alex (Done): This formulation in the prompt looks strange. Why is it used? attribute:  attribute-value pairs. 
%Todo Alex (Done): "All parts of the product title must be assigned to an attribute" Did you also test "should"? Or removing this sentense, because it sounds like to force the model to output stuff even if it is very unsure.

% \begin{figure}[ht]
% \includegraphics[width=.45\textwidth]{figures/closed_information_extraction.PNG}
% \caption{Two zero-shot prompt designs for closed product information extraction using ChatGPT. The example prompts consist of a task description and a task input.}
% \label{fig:closed_extraction_prompt}
% \end{figure}

% \begin{figure}[ht]
% \includegraphics[width=.45\textwidth]{figures/generated_knowledge.PNG}
% \caption{Task description and task input of an example prompt that first generates knowledge about the product and afterwards extracts an explicit attribute value. \rs{Should this figure contain the output as well?}}
% \label{fig:generated_prompt}
% \end{figure}

\vspace{.2cm}\noindent\textbf{Open Product Information Extraction.}
The goal of our open prompts is to extract all attribute-value pairs from a product title. 
%These advantes are vage and their utility depends on the use case. Better remove. It has two main advantages over closed extraction. First, it is open to any available product attribute and does ask for a specific attribute. Second, only one extraction is necessary to extract multiple attribute/value pairs. 
%This approach does not ask for specific attribute/value pairs and lets the model identify the relevant attribute/value pairs. 
%Remove this sentense as too vague for scientific paper. What means works fine? WHich use case has these requirements? If 
Prompt (d) in Figure \ref{fig:extraction_prompt_new} shows an example prompt of open extraction. We explicitly describe the output format so that  attribute/value pairs can be parsed from the response and compared to the ground truth in the test set.

% \begin{figure}[ht]
% \includegraphics[width=.45\textwidth]{figures/open_information_extraction.PNG}
% \caption{Task description, task input and task output of an example prompt that uses open information to extract all attribute-value pairs from the product title.}
% \label{fig:open_extraction_prompt}
% \end{figure}

\vspace{.2cm}\noindent\textbf{Results.}
Table \ref{tab:zero_shot_results} shows the results of all zero-shot prompts and baselines.
Overall, our results show that ChatGPT achieves %results 
comparable performance to that of NER.
AVEQA shows an exceptional performance with almost perfect Precision, Recall and F1 %being close to 1 
and clearly outperforms the zero-shot results with ChatGPT.
% Todo Alex (Done): Be presice, you talk about Zero-shot case, right? So the term "additoinal training data" is misleading. Please do not use it.
However, both baselines (AVEQA and NER) are finetuned on 71,266 attribute/value pairs and require additional computing for fine-tuning whereas ChatGPT needs only an explicit prompt.
Requiring no training data for task-specific finetuning to achieve these results is an advantage of ChatGPT over AVEQA and NER.

The explicit prompt (Explicit) with a request for an attribute, a short answer and telling the model how to respond (recall prompt example in Figure~\ref{fig:extraction_prompt_new}(b)) achieves an F1 score of 87.86\%.
Specifically, telling the model how to respond if the attribute is unknown to the model, highly impacts the model's precision. Without 'unk.' the model's precision drops by 24.79\%.
%This zero-shot performance of ChatGPT is possible if the prompt explicitly requests a specific attribute and tells the model what to do if the target attribute value is not contained in the product title. \rs{Maybe we can be further explicit and talk about the addition, e.g., "First the addition of an explicit request (Req.) boosts the performance significantly (+41.61 F1)..."} Without the `N/A' addition to the prompts, ChatGPT responds with different outputs that are semantically similar to `I do not know.' but the outputs cannot be simply parsed. This explains why the precision drops significantly if the `N/A' addition is not provided.
Generating knowledge about the product improves the results only marginally in our experiments and since the task descriptions are longer, costs more per Title.

The open extraction is comparable to the best closed extraction results with respect to precision but the recall is more than 40\% lower. ChatGPT has no guidance on how the attributes are named and, thus, proposes its own attribute names. Since we check for an exact match on the attribute names during the evaluation, the unknown attribute names generated by ChatGPT are not found. Prompt (d) in Figure \ref{fig:extraction_prompt_new} illustrates this behaviour for the attribute \emph{"Resolution"}, which is extracted as \emph{"Megapixels"} by ChatGPT. A possible solution to this issue is to use demonstrations with an example schema, which ChatGPT can use to build a schema for the extraction. We explore this idea in Section \ref{sec:in_context_learning}.

We noticed that ChatGPT normalizes attribute values for both open and closed extraction. Table \ref{tab:value_normalization} shows examples of how ChatGPT normalizes attribute values.
%During the result inspection, we observe that ChatGPT interprets and normalizes the extracted attribute values. An example of this behaviour is the attribute \emph{"Number of Processor Cores"} of the category \emph{"Laptops"}. In the MAVE dataset, the target attribute value is usually "dual-core" or "quad-core". ChatGPT seems to extract and interpret these values because it returns "2" for "dual-core" and "4" for "quad-core". Since the attribute "Number of Processor Cores" explicitly asks for a number, we added "2" and "4" as accepted attribute values to the ground truth.

\begin{table}[]
\setlength{\tabcolsep}{4pt}
\caption{Precision (P), Recall (R), F1, $\Delta$ F1 and Cost (\textcent) per Title of the Zero-Shot prompts and fully trained baselines.}
\label{tab:zero_shot_results}
\scalebox{0.95}{\begin{tabular}{@{}l|ccccc@{}}
\toprule
Configuration         & P & R & F1 & $\Delta F1$ over & Cost (\textcent) \\ 
         & & & & Simple & per Title \\ \midrule
Simple             & 30.60                  & 39.72                 & 34.57                  & -        & 0.0175                                     \\
Explicit    & \textbf{87.96}                 & 87.76                 & 87.86                  & +53.29                        & 0.0196                                     \\
Explicit w/o sh.,unk.              & 67.44                 & 87.53                 & 76.18                  & +41.61                        & \textbf{0.0161}                                    \\
Explicit w/o unk.        & 63.17                 & 81.99                 & 71.36                  & +36.79                        & 0.0176                                     \\

Gen. Knowledge & 87.79                 & \textbf{87.99}                & \textbf{87.89}                  & +53.32                        & 0.0242                                     \\
Open                  & 70.27                 & 36.45                 & 48.00                     & +13.43                        & 0.0336                                     \\\midrule
NER (fully trained)           & 90.80                  & 84.3                 & 87.43                  &        \multicolumn{1}{c}{-}  & \multicolumn{1}{c}{-}   \\
AVEQA (fully trained) & \textbf{99.52} & \textbf{99.28} & \textbf{99.41} &\multicolumn{1}{c}{-} &\multicolumn{1}{c}{-} 
\\ \bottomrule
\end{tabular}}
\end{table}

\section{In-Context Learning}
\label{sec:in_context_learning}

In-context learning is a paradigm that allows language models to learn a task based on a few example inputs and outputs also called demonstrations~\cite{brown_language_2020,dong_survey_2023}. 
In the context of this paper, we combine the closed prompt `Generated Knowledge` and the open prompt with demonstrations to analyze the impact of demonstrations on the performance of ChatGPT for Product Information Extraction. First, %ly, 
we explain how the demonstrations are selected and combined with the prompts. %Secondly, 
Then, we present the results of our experiments.

\vspace{.2cm}\noindent\textbf{Demonstration Selection.}
The source for the used demonstrations is the reduced training set. Demonstrations are useful if they are semantically similar to the final task~\cite{liu_what_2022}. Since we know the product category of the input product title and assume that product offers with the same product category are semantically similar, the selected demonstrations are drawn from the same product category as the product offer for which we run the product information extraction. Both the closed and the open prompt are combined with one, three and six demonstrations. For the prompts with three and six demonstrations, two-thirds of the demonstrations are product offers where the value of the attribute/value pair is contained in the title and one-third are product offers where the value of the attribute/value pair is not contained in the title. This selection of demonstrations covers different possible structures of the task~\cite{levy_diverse_2022}. All attribute values of the demonstrations are manually checked and normalized like the examples listed in Table \ref{tab:value_normalization}.
%similar to the ground truth attribute values such as \emph{"8 hour"} and \emph{"13.3-inch"} which are normalized to \emph{"8 hours"} and \emph{"13.3 inches"}, respectively.
%Like in the ground truth attribute values like \emph{``8 hour``} and \emph{``13.3-inch``} are normalized to \emph{``8 hours``} and \emph{``13.3 inches``}.

\begin{figure}[ht]
\includegraphics[width=.475\textwidth]{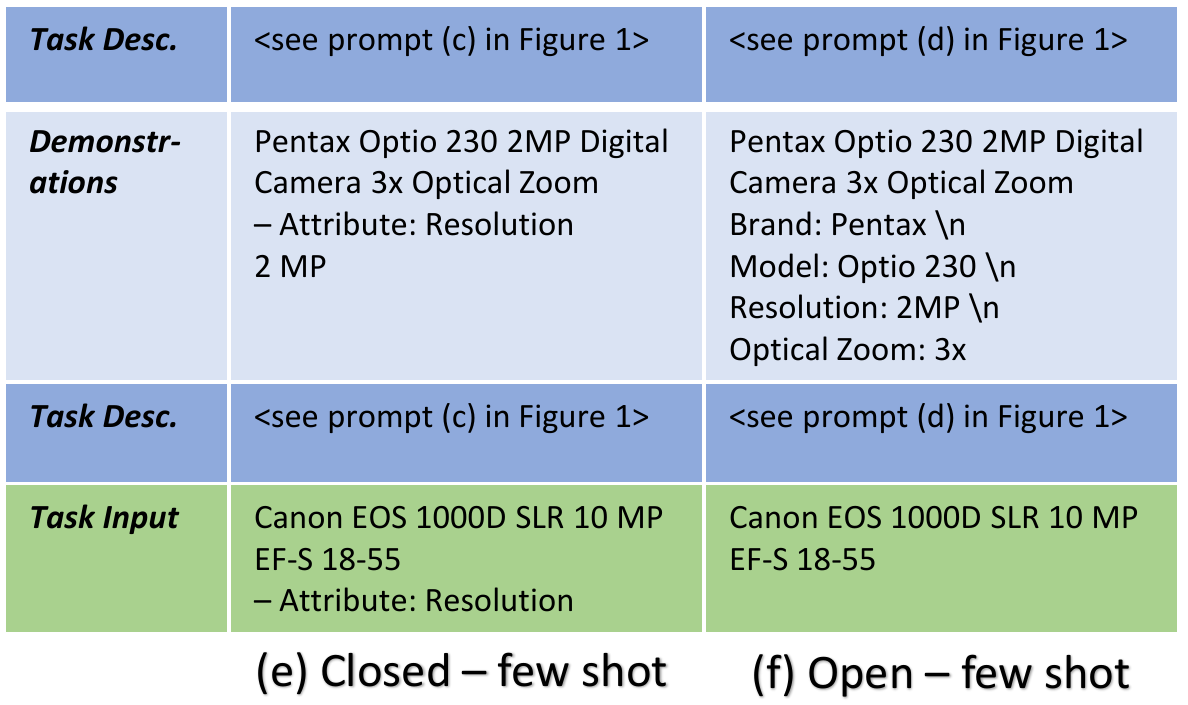}
\caption{This example prompt utilizes demonstrations to guide the open product information extraction.}
\label{fig:few_shot_cropped}
\end{figure}

\vspace{.2cm}\noindent\textbf{Combination of Prompts and Demonstrations.} For the in-context learning prompts, we first provide the task description and add the task demonstrations. Each task demonstration consists of an example input and an example output. Afterwards, we repeat the task description and add the task input.
The prompt examples (e) and (f) in Figure \ref{fig:few_shot_cropped} illustrate closed and open prompts with demonstrations.

%\begin{figure}[ht]
%\includegraphics[width=.45\textwidth]{figures/gk_few_shots_cropped_V2.pdf}
%\caption{This example prompt utilizes demonstrations to guide the closed product information extraction.}
%\label{fig:gk_few_shots_cropped}
%\end{figure}

%\begin{figure}[ht]
%\includegraphics[width=.45\textwidth]{figures/open_few_shots_cropped_V2.pdf}
%\caption{This example prompt utilizes demonstrations to guide the open product information extraction.}
%\label{fig:open_few_shots_cropped}
%\end{figure}

\begin{table}[]
\setlength{\tabcolsep}{4pt}
\caption{Precision (P), Recall (R), F1, $\Delta$ F1 and Cost (\textcent) per Title of the prompts with one, three and six demonstrations (Shots) and fully trained baselines}
\label{tab:few_shot_results}
\scalebox{0.925}{\begin{tabular}{@{}ll|ccccc@{}}
\toprule
Config. & \# Shots & P & R & F1 & $\Delta$F1 over & Cost (\textcent)                   \\ 
 &  &  &  &  & zero-shot  &  per Title                    \\ \midrule
Closed        & Zero  & 87.79                 & 87.99                 & 87.89                  & \multicolumn{1}{c}{-}        & \textbf{0.0242}                     \\
Closed        & One   & 90.99                 & 93.30                  & 92.13                  & +4.24                         & 0.0482                     \\
Closed        & Three & 91.17                 & 95.38                 & 93.23                  & +5.34                         & 0.0705                     \\
Closed        & Six   & 93.17                 & \textbf{95.94}                 & \textbf{94.51}                  & +6.62                         & 0.1037                     \\ \midrule
Open          & Zero  & 86.92                 & 43.46                 & 57.95                  & -29.94                       & 0.0305                     \\
Open          & One   & \textbf{95.65}                 & 86.37                 & 90.77                  & +2.88                         & 0.0679                     \\
Open          & Three & 91.08                 & 81.03                 & 83.92                  & -3.97                        & 0.0910                     \\
Open          & Six   & 91.48                 & 82.77                 & 85.73                  & -2.16                        & 0.1203   \\ \midrule
\multicolumn{2}{l|}{NER (fully trained)}      & 90.80                  & 84.3                 & 87.43                  &        \multicolumn{1}{c}{-}  & \multicolumn{1}{c}{-}   \\
\multicolumn{2}{l|}{AVEQA (fully trained)} & \textbf{99.52} & \textbf{99.28} & \textbf{99.41} &\multicolumn{1}{c}{-} &\multicolumn{1}{c}{-} 
\\ \bottomrule
\end{tabular}}
\end{table}

\vspace{.2cm}\noindent\textbf{Results.}
The results with in-context learning in Table \ref{tab:few_shot_results} show that demonstrations improve the extraction of attribute/value pairs of closed and open prompts. A single example (one-shot) is already sufficient to increase the closed extraction results by 4.24\% in terms of F1. Additional demonstrations add another 1-2\% in F1, with a maximal F1 of 94.51 achieved with (only) six samples.

The open extraction benefits the most from the demonstrations. A single example is sufficient to increase the performance of ChatGPT by almost 33\% in terms of F1. We noticed that the model picks up on the structure of the demonstrations and uses the demonstration attribute/value pairs to extract the attribute/value pairs from the task input. Looking at the example prompt (f) in Figure \ref{fig:few_shot_cropped}, for example, the model learns from the task demonstration that the attribute value \emph{"10 MP"} has to be extracted as \emph{"Resolution: 10 MP"}, rather than \emph{"Megapixels: 10 MP"} as it can happen if no demonstration is provided.
%ChatGPT's F1 increases if an open prompt with one demonstration is used but this comes at the cost of a high loss in recall compared to closed extraction.
In our experiments, open extraction prompts with three and six demonstrations resulted in a worse performance than experiments with only one demonstration. We assume that this behaviour is caused by the longer prompts but additional experiments are necessary to verify this hypothesis.

\vspace{.2cm}\noindent\textbf{Cost of the extraction.}
%\rs{What do you think about explicitly expressing here the trade-off between high-cost and high-quality and the way they are expressed in open/closed extraction? I think it can make the final paragraph more general; yet, still expressing what you wanted in terms of the cost.} 
On the one hand, the closed extraction few-shot prompts are 25\% cheaper per API call and exhibit an 8\% higher F1 recall than the open extraction few-shot prompts. On the other hand, open extraction prompts can extract multiple attributes, which makes them cheaper than closed extraction prompts if two or more attributes have to be extracted. 

%Todo Alex (Done): Shorten list of authors in the references using "et. al."
%\input{sections/related_work.tex}
\section{Conclusion}
\label{sec:conclusion}

In this paper, we systematically evaluate ChatGPT(gpt3.5-turbo-0301) for the task of product information extraction. We experiment with different zero-shot prompt designs, which show that a proper prompt design is sufficient to achieve similar performance to that of a fine-tuned NER baseline.
%Todo Alex: Change conclusion to not focus on specific performance improvement, but to state that a similar performance is reached with much less training examples.
By adding demonstrations to the prompts, the results are further improved.
%we are able to further improve F1 score of 94.51\%.
While state-of-the-art models like AVEQA outperform ChatGPT's attribute/value pair extractions, they require far more training examples (71,266 vs. 6) and computation ($\sim$10 hrs.) for fine-tuning.
Our experiments also show that ChatGPT can normalize the extracted attribute/value pairs. These normalizations improve the quality of the extracted attribute/value pairs.

\bibliographystyle{ACM-Reference-Format}
\bibliography{library}


\begin{thebibliography}{22}

%%% the Web address.

\ifx \showCODEN    \undefined \def \showCODEN     #1{\unskip}     \fi
\ifx \showDOI      \undefined \def \showDOI       #1{#1}\fi
\ifx \showISBNx    \undefined \def \showISBNx     #1{\unskip}     \fi
\ifx \showISBNxiii \undefined \def \showISBNxiii  #1{\unskip}     \fi
\ifx \showISSN     \undefined \def \showISSN      #1{\unskip}     \fi
\ifx \showLCCN     \undefined \def \showLCCN      #1{\unskip}     \fi
\ifx \shownote     \undefined \def \shownote      #1{#1}          \fi
\ifx \showarticletitle \undefined \def \showarticletitle #1{#1}   \fi
\ifx \showURL      \undefined \def \showURL       {\relax}        \fi
% invisible to TeX
\providecommand\bibfield[2]{#2}
\providecommand\bibinfo[2]{#2}
\providecommand\natexlab[1]{#1}
\providecommand\showeprint[2][]{arXiv:#2}

\bibitem[Brown et~al\mbox{.}(2020)]%
        {brown_language_2020}
\bibfield{author}{\bibinfo{person}{Tom~B. Brown}, \bibinfo{person}{Benjamin
  Mann}, \bibinfo{person}{Nick Ryder}, {et~al\mbox{.}}}
  \bibinfo{year}{2020}\natexlab{}.
\newblock \showarticletitle{Language {Models} are {Few}-{Shot} {Learners}}.
\newblock \bibinfo{journal}{\emph{arXiv:2005.14165 [cs]}} (\bibinfo{date}{July}
  \bibinfo{year}{2020}).
\newblock
\newblock
\shownote{arXiv: 2005.14165}.


\bibitem[Chowdhery et~al\mbox{.}(2022)]%
        {chowdhery_palm_2022}
\bibfield{author}{\bibinfo{person}{Aakanksha Chowdhery},
  \bibinfo{person}{Sharan Narang}, \bibinfo{person}{Jacob Devlin},
  {et~al\mbox{.}}} \bibinfo{year}{2022}\natexlab{}.
\newblock \bibinfo{title}{{PaLM}: {Scaling} {Language} {Modeling} with
  {Pathways}}.
\newblock
\newblock
\urldef\tempurl%
\url{https://doi.org/10.48550/arXiv.2204.02311}
\showDOI{\tempurl}
\newblock
\shownote{arXiv:2204.02311 [cs]}.


\bibitem[Dong et~al\mbox{.}(2023)]%
        {dong_survey_2023}
\bibfield{author}{\bibinfo{person}{Qingxiu Dong}, \bibinfo{person}{Lei Li},
  \bibinfo{person}{Damai Dai}, {et~al\mbox{.}}}
  \bibinfo{year}{2023}\natexlab{}.
\newblock \bibinfo{title}{A {Survey} on {In}-context {Learning}}.
\newblock
\newblock
\newblock
\shownote{arXiv:2301.00234 [cs]}.


\bibitem[He et~al\mbox{.}(2021)]%
        {he_deberta_2021}
\bibfield{author}{\bibinfo{person}{Pengcheng He}, \bibinfo{person}{Xiaodong
  Liu}, \bibinfo{person}{Jianfeng Gao}, {et~al\mbox{.}}}
  \bibinfo{year}{2021}\natexlab{}.
\newblock \showarticletitle{{DeBERTa}: {Decoding}-enhanced {BERT} with
  {Disentangled} {Attention}}.
\newblock \bibinfo{journal}{\emph{arXiv:2006.03654 [cs]}} (\bibinfo{date}{Jan.}
  \bibinfo{year}{2021}).
\newblock
\newblock
\shownote{arXiv: 2006.03654}.


\bibitem[Levy et~al\mbox{.}(2022)]%
        {levy_diverse_2022}
\bibfield{author}{\bibinfo{person}{Itay Levy}, \bibinfo{person}{Ben Bogin},
  {and} \bibinfo{person}{Jonathan Berant}.} \bibinfo{year}{2022}\natexlab{}.
\newblock \bibinfo{title}{Diverse {Demonstrations} {Improve} {In}-context
  {Compositional} {Generalization}}.
\newblock
\newblock
\newblock
\shownote{arXiv:2212.06800 [cs]}.


\bibitem[Liu et~al\mbox{.}(2022a)]%
        {liu_generated_2022}
\bibfield{author}{\bibinfo{person}{Jiacheng Liu}, \bibinfo{person}{Alisa Liu},
  \bibinfo{person}{Ximing Lu}, {et~al\mbox{.}}}
  \bibinfo{year}{2022}\natexlab{a}.
\newblock \showarticletitle{Generated {Knowledge} {Prompting} for {Commonsense}
  {Reasoning}}. In \bibinfo{booktitle}{\emph{{ACL2022}}}.
  \bibinfo{pages}{3154--3169}.
\newblock


\bibitem[Liu et~al\mbox{.}(2022b)]%
        {liu_what_2022}
\bibfield{author}{\bibinfo{person}{Jiachang Liu}, \bibinfo{person}{Dinghan
  Shen}, \bibinfo{person}{Yizhe Zhang}, {et~al\mbox{.}}}
  \bibinfo{year}{2022}\natexlab{b}.
\newblock \showarticletitle{What {Makes} {Good} {In}-{Context} {Examples} for
  {GPT}-3?}. In \bibinfo{booktitle}{\emph{{DeeLIO2022}}}.
  \bibinfo{pages}{100--114}.
\newblock


\bibitem[Narayan et~al\mbox{.}(2022)]%
        {narayan_can_2022}
\bibfield{author}{\bibinfo{person}{Avanika Narayan}, \bibinfo{person}{Ines
  Chami}, \bibinfo{person}{Laurel Orr}, {et~al\mbox{.}}}
  \bibinfo{year}{2022}\natexlab{}.
\newblock \showarticletitle{Can {Foundation} {Models} {Wrangle} {Your}
  {Data}?}. In \bibinfo{booktitle}{\emph{{VLDB2022}}}
  \emph{(\bibinfo{series}{4}, Vol.~\bibinfo{volume}{16})}.
  \bibinfo{pages}{738--746}.
\newblock


\bibitem[Ni et~al\mbox{.}(2019)]%
        {ni_justifying_2019}
\bibfield{author}{\bibinfo{person}{Jianmo Ni}, \bibinfo{person}{Jiacheng Li},
  {and} \bibinfo{person}{Julian McAuley}.} \bibinfo{year}{2019}\natexlab{}.
\newblock \showarticletitle{Justifying {Recommendations} using
  {Distantly}-{Labeled} {Reviews} and {Fine}-{Grained} {Aspects}}. In
  \bibinfo{booktitle}{\emph{{EMNLP2019}}}. \bibinfo{pages}{188--197}.
\newblock


\bibitem[Ouyang et~al\mbox{.}(2022)]%
        {ouyang_training_2022}
\bibfield{author}{\bibinfo{person}{Long Ouyang}, \bibinfo{person}{Jeffrey Wu},
  \bibinfo{person}{Xu Jiang}, {et~al\mbox{.}}} \bibinfo{year}{2022}\natexlab{}.
\newblock \showarticletitle{Training language models to follow instructions
  with human feedback}. In \bibinfo{booktitle}{\emph{{NeurIPS2022}}},
  Vol.~\bibinfo{volume}{35}. \bibinfo{pages}{27730--27744}.
\newblock


\bibitem[Putthividhya and Hu(2011)]%
        {putthividhya_bootstrapped_2011}
\bibfield{author}{\bibinfo{person}{Duangmanee Putthividhya} {and}
  \bibinfo{person}{Junling Hu}.} \bibinfo{year}{2011}\natexlab{}.
\newblock \showarticletitle{Bootstrapped {Named} {Entity} {Recognition} for
  {Product} {Attribute} {Extraction}}. In
  \bibinfo{booktitle}{\emph{{EMNLP2011}}}. \bibinfo{pages}{1557--1567}.
\newblock


\bibitem[Wang et~al\mbox{.}(2020)]%
        {wang_learning_2020}
\bibfield{author}{\bibinfo{person}{Qifan Wang}, \bibinfo{person}{Li Yang},
  \bibinfo{person}{Bhargav Kanagal}, {et~al\mbox{.}}}
  \bibinfo{year}{2020}\natexlab{}.
\newblock \showarticletitle{Learning to {Extract} {Attribute} {Value} from
  {Product} via {Question} {Answering}: {A} {Multi}-task {Approach}}. In
  \bibinfo{booktitle}{\emph{{KDD2020}}}. \bibinfo{pages}{47--55}.
\newblock


\bibitem[Wei et~al\mbox{.}(2013)]%
        {wei_survey_2013}
\bibfield{author}{\bibinfo{person}{Bifan Wei}, \bibinfo{person}{Jun Liu},
  \bibinfo{person}{Qinghua Zheng}, \bibinfo{person}{Wei Zhang},
  \bibinfo{person}{Xiaoyu Fu}, {and} \bibinfo{person}{Boqin Feng}.}
  \bibinfo{year}{2013}\natexlab{}.
\newblock \showarticletitle{A {SURVEY} {OF} {FACETED} {SEARCH}}.
\newblock \bibinfo{journal}{\emph{Journal of Web Engineering}}
  (\bibinfo{date}{Nov.} \bibinfo{year}{2013}), \bibinfo{pages}{041--064}.
\newblock


\bibitem[Wei et~al\mbox{.}(2022)]%
        {wei_emergent_2022}
\bibfield{author}{\bibinfo{person}{Jason Wei}, \bibinfo{person}{Yi Tay},
  \bibinfo{person}{Rishi Bommasani}, {et~al\mbox{.}}}
  \bibinfo{year}{2022}\natexlab{}.
\newblock \bibinfo{title}{Emergent {Abilities} of {Large} {Language} {Models}}.
\newblock
\newblock
\newblock
\shownote{arXiv:2206.07682 [cs]}.


\bibitem[Xu et~al\mbox{.}(2019)]%
        {xu_scaling_2019}
\bibfield{author}{\bibinfo{person}{Huimin Xu}, \bibinfo{person}{Wenting Wang},
  \bibinfo{person}{Xin Mao}, {et~al\mbox{.}}} \bibinfo{year}{2019}\natexlab{}.
\newblock \showarticletitle{Scaling up {Open} {Tagging} from {Tens} to
  {Thousands}: {Comprehension} {Empowered} {Attribute} {Value} {Extraction}
  from {Product} {Title}}. In \bibinfo{booktitle}{\emph{{ACL2019}}}.
  \bibinfo{pages}{5214--5223}.
\newblock


\bibitem[Yan et~al\mbox{.}(2021)]%
        {yan_adatag_2021}
\bibfield{author}{\bibinfo{person}{Jun Yan}, \bibinfo{person}{Nasser Zalmout},
  \bibinfo{person}{Yan Liang}, {et~al\mbox{.}}}
  \bibinfo{year}{2021}\natexlab{}.
\newblock \showarticletitle{{AdaTag}: {Multi}-{Attribute} {Value} {Extraction}
  from {Product} {Profiles} with {Adaptive} {Decoding}}. In
  \bibinfo{booktitle}{\emph{{ACL2021}}}. \bibinfo{pages}{4694--4705}.
\newblock


\bibitem[Yang et~al\mbox{.}(2022)]%
        {yang_mave_2022}
\bibfield{author}{\bibinfo{person}{Li Yang}, \bibinfo{person}{Qifan Wang},
  \bibinfo{person}{Zac Yu}, {et~al\mbox{.}}} \bibinfo{year}{2022}\natexlab{}.
\newblock \showarticletitle{{MAVE}: {A} {Product} {Dataset} for {Multi}-source
  {Attribute} {Value} {Extraction}}. In \bibinfo{booktitle}{\emph{{WSDM2022}}}.
  \bibinfo{publisher}{Association for Computing Machinery},
  \bibinfo{address}{New York, NY, USA}, \bibinfo{pages}{1256--1265}.
\newblock


\bibitem[Yong et~al\mbox{.}(2023)]%
        {yong_bloom1_2023}
\bibfield{author}{\bibinfo{person}{Zheng-Xin Yong}, \bibinfo{person}{Hailey
  Schoelkopf}, \bibinfo{person}{Niklas Muennighoff}, {et~al\mbox{.}}}
  \bibinfo{year}{2023}\natexlab{}.
\newblock \bibinfo{title}{{BLOOM}+1: {Adding} {Language} {Support} to {BLOOM}
  for {Zero}-{Shot} {Prompting}}.
\newblock
\newblock
\newblock
\shownote{arXiv:2212.09535 [cs]}.


\bibitem[Zhang et~al\mbox{.}(2022)]%
        {zhang_oa-mine_2022}
\bibfield{author}{\bibinfo{person}{Xinyang Zhang}, \bibinfo{person}{Chenwei
  Zhang}, \bibinfo{person}{Xian Li}, {et~al\mbox{.}}}
  \bibinfo{year}{2022}\natexlab{}.
\newblock \showarticletitle{{OA}-{Mine}: {Open}-{World} {Attribute} {Mining}
  for {E}-{Commerce} {Products} with {Weak} {Supervision}}. In
  \bibinfo{booktitle}{\emph{{WWW2022}}}. \bibinfo{pages}{3153--3161}.
\newblock


\bibitem[Zhao et~al\mbox{.}(2021)]%
        {zhao_calibrate_2021}
\bibfield{author}{\bibinfo{person}{Zihao Zhao}, \bibinfo{person}{Eric Wallace},
  \bibinfo{person}{Shi Feng}, {et~al\mbox{.}}} \bibinfo{year}{2021}\natexlab{}.
\newblock \showarticletitle{Calibrate {Before} {Use}: {Improving} {Few}-shot
  {Performance} of {Language} {Models}}. In
  \bibinfo{booktitle}{\emph{{ICML2021}}}. \bibinfo{pages}{12697--12706}.
\newblock


\bibitem[Zheng et~al\mbox{.}(2018)]%
        {zheng_opentag_2018}
\bibfield{author}{\bibinfo{person}{Guineng Zheng}, \bibinfo{person}{Subhabrata
  Mukherjee}, \bibinfo{person}{Xin~Luna Dong}, {et~al\mbox{.}}}
  \bibinfo{year}{2018}\natexlab{}.
\newblock \showarticletitle{{OpenTag}: {Open} {Attribute} {Value} {Extraction}
  from {Product} {Profiles}}. In \bibinfo{booktitle}{\emph{{KDD2018}}}.
  \bibinfo{pages}{1049--1058}.
\newblock


\bibitem[Zhu et~al\mbox{.}(2020)]%
        {zhu_multimodal_2020}
\bibfield{author}{\bibinfo{person}{Tiangang Zhu}, \bibinfo{person}{Yue Wang},
  \bibinfo{person}{Haoran Li}, {et~al\mbox{.}}}
  \bibinfo{year}{2020}\natexlab{}.
\newblock \showarticletitle{Multimodal {Joint} {Attribute} {Prediction} and
  {Value} {Extraction} for {E}-commerce {Product}}. In
  \bibinfo{booktitle}{\emph{{EMNLP2020}}}. \bibinfo{pages}{2129--2139}.
\newblock


\end{thebibliography}

\end{document}